\documentclass{article} 
\usepackage{iclr2026_conference,times}

\usepackage{amsmath,amsfonts,bm}

\def\eqref#1{equation~\ref{#1}}

\def\1{\bm{1}}

\def\ra{{\textnormal{a}}}

\def\rx{{\textnormal{x}}}

\def\rva{{\mathbf{a}}}

\def\erva{{\textnormal{a}}}

\def\ervx{{\textnormal{x}}}

\def\rmA{{\mathbf{A}}}

\def\vmu{{\bm{\mu}}}
\def\vtheta{{\bm{\theta}}}
\def\va{{\bm{a}}}

\def\ve{{\bm{e}}}

\def\vx{{\bm{x}}}

\def\eva{{a}}

\def\mA{{\bm{A}}}

\def\mH{{\bm{H}}}
\def\mI{{\bm{I}}}
\def\mJ{{\bm{J}}}

\def\mX{{\bm{X}}}

\def\mSigma{{\bm{\Sigma}}}

\DeclareMathAlphabet{\mathsfit}{\encodingdefault}{\sfdefault}{m}{sl}
\SetMathAlphabet{\mathsfit}{bold}{\encodingdefault}{\sfdefault}{bx}{n}
\newcommand{\tens}[1]{\bm{\mathsfit{#1}}}
\def\tA{{\tens{A}}}

\def\tX{{\tens{X}}}

\def\gG{{\mathcal{G}}}

\def\sA{{\mathbb{A}}}
\def\sB{{\mathbb{B}}}

\def\sS{{\mathbb{S}}}

\def\emA{{A}}

\newcommand{\etens}[1]{\mathsfit{#1}}

\def\etA{{\etens{A}}}

\newcommand{\E}{\mathbb{E}}

\newcommand{\R}{\mathbb{R}}

\newcommand{\KL}{D_{\mathrm{KL}}}
\newcommand{\Var}{\mathrm{Var}}

\newcommand{\Cov}{\mathrm{Cov}}

\newcommand{\normltwo}{L^2}
\newcommand{\normlp}{L^p}

\newcommand{\parents}{Pa} 

\usepackage{hyperref}
\usepackage{url}

\usepackage{booktabs,multirow}
\usepackage{tabularx}

\usepackage{amsmath}
\usepackage{stackengine}

\usepackage{graphicx} 

\usepackage{hyperref} 
\hypersetup{
    colorlinks=true,   
    urlcolor=blue,     
    linkcolor=red,     
    citecolor=green    
}

\title{Domain Expansion: A Latent Space Construction Framework for Multi-Task Learning}

\author{Chi-Yao Huang\textsuperscript{*, 1}, Khoa Vo\textsuperscript{*, 1}, Aayush Atul Verma\textsuperscript{*, 1}, Duo Lu\textsuperscript{2}, Yezhou Yang\textsuperscript{1} \\
\textsuperscript{1}Arizona State University \quad
\textsuperscript{2}Rider University \\
\texttt{\{cy.huang, ngocbach, averma90, yz.yang\}@asu.edu} \\
\texttt{\{dlu@rider.edu\}} \\
\small{\textsuperscript{*} Equal contribution} 
}

\iclrfinalcopy 
\begin{document}

\maketitle

\begin{abstract}
Training a single network with multiple objectives often leads to conflicting gradients that degrade shared representations, forcing them into a compromised state that is suboptimal for any single task—a problem we term latent representation collapse. We introduce Domain Expansion, a framework that prevents these conflicts by restructuring the latent space itself. Our framework uses a novel orthogonal pooling mechanism to construct a latent space where each objective is assigned to a mutually orthogonal subspace. We validate our approach across diverse benchmarks—including ShapeNet, MPIIGaze, and Rotated MNIST—on challenging multi-objective problems combining classification with pose and gaze estimation. Our experiments demonstrate that this structure not only prevents collapse but also yields an explicit, interpretable, and compositional latent space where concepts can be directly manipulated. 
\end{abstract}
    
\begin{figure}[!b]
  \centering
  \includegraphics[width=0.85\linewidth]{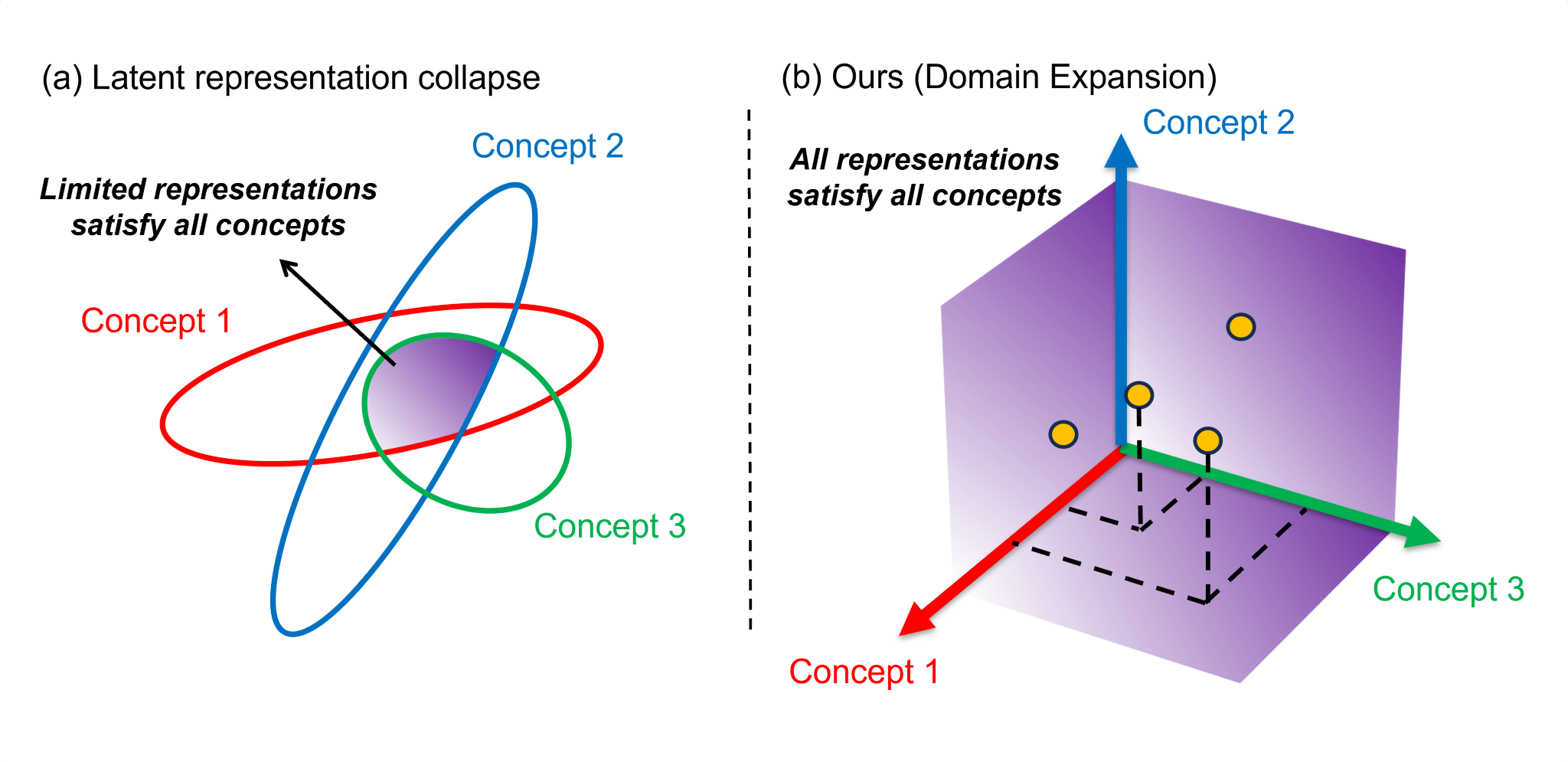}
  \caption{
    \textbf{(a) Latent representation collapse.} In standard multi-task learning, competing objectives lead to latent representation collapse, where the solution spaces for different concepts (colored ellipses) overlap in only a small, compromised region.
    \textbf{(b) Domain Expansion.} In contrast, our method assigns each concept to an orthogonal basis vector in the latent space, preventing interference and creating a structured, interpretable representation where features for each concept are clearly separated.
    }
  \label{main}
\end{figure}

\section{Introduction}
Representation learning seeks to map raw data into meaningful latent features in latent space, a principle that underpins successes across machine learning, from classification \citep{SimCLR, SupCon} to large-scale multimodal learning \citep{CLIP, LLava}. A common paradigm is to train a single, unified network to satisfy multiple learning objectives simultaneously—for example, performing both classification and regression on the same input. However, this approach exposes a fundamental challenge: competing objectives often produce conflicting gradients, pulling shared latent features in opposing directions. This interference can degrade model performance, a phenomenon we formalize as \emph{latent representation collapse} (Fig.~\ref{main}(a)).

When collapse occurs, the network carves out only a small, compromised region of the latent space that partially satisfies all objectives, failing to excel at any. This not only limits predictive accuracy but also leads to entangled, uninterpretable representations where the underlying factors of variation are obscured. While existing multi-task learning methods attempt to mitigate this by re-weighting task losses or projecting gradients at runtime \citep{PCGrad, MGDA}, these approaches act on the optimization process itself, not the structure of the latent space. The core problem of designing a representation space that is inherently robust to interference remains underexplored.

To address this gap, we propose Domain Expansion, a new framework for constructing latent spaces that are multi-objective by design. Instead of mediating gradient conflicts, we prevent them structurally. The core of our framework is orthogonal pooling, a lightweight architectural primitive that constrains features for different objectives to lie in mutually orthogonal subspaces. This enforcement of orthogonality ensures that learning one objective cannot interfere with another, allowing the model to learn potent, disentangled features for all tasks simultaneously (Fig.~\ref{main}(b)).

We validate Domain Expansion on a diverse set of benchmarks, including ShapeNet \citep{shapenet}, MPIIGaze \citep{MPIIGaze}, and rotated MNIST \citep{MNIST}, tackling multi-objective problems that combine classification with various forms of regression like pose and gaze estimation. Across all experiments, our method successfully resolves latent representation collapse. Furthermore, we demonstrate that the resulting latent space is not a black box; it possesses an explicit, interpretable structure where distinct objectives are disentangled by design, enabling analysis of their relationships. Our contributions are:

\begin{itemize}
\item We formalize latent representation collapse, a critical failure mode in multi-objective representation learning.
\item We introduce Domain Expansion, a framework that uses orthogonal pooling to construct a latent space with mutually orthogonal subspaces, preventing task interference by design.
\item We demonstrate that our method constructs an explicit and interpretable latent space, where orthogonal axes correspond to distinct concepts, enabling compositional inference and analysis.
\end{itemize}

\section{Related Work}

The goal of representation learning is to build latent spaces that capture meaningful factors of variation from raw data. Modern methods often learn such representations by optimizing for multiple, sometimes implicit, objectives. For instance, contrastive learning \citep{SimCLR, MoCo, SupCon} structures the latent space by pulling similar samples together while pushing dissimilar ones apart. Large multimodal models like CLIP \citep{CLIP} learn powerful, transferable features by aligning representations from different modalities, such as images and text. The success of these models hinges on their ability to create a unified representation that satisfies these diverse learning signals.

This principle is formalized in multi-task learning (MTL), which aims to improve generalization and efficiency by training a single model on multiple tasks simultaneously \citep{taskonomy}. However, a central challenge in MTL is negative transfer \citep{negative_transfer}, where optimizing for one task degrades the performance of another. This issue often arises from conflicting gradients, where different task-specific losses pull the shared network parameters in opposing directions. This conflict can lead to the phenomenon we term latent representation collapse, where the learned features are a poor compromise that fails to adequately solve any single task.

A dominant line of work addresses this challenge by manipulating task gradients during the optimization process. These methods aim to find a less conflicting update vector by altering the gradients from individual tasks. For example, GradNorm \citep{GradNorm} dynamically adjusts the weights of each task's loss to balance their training rates. Other methods focus on the geometry of the gradients themselves. PCGrad \citep{PCGrad} and IMTL \citep{IMTL} identify conflicting gradients and projects them onto the normal plane of others to remove the conflicting components. Building on this, methods like CAGrad \citep{CAGrad} and MGDA \citep{MGDA} seek a common gradient that represents a Pareto optimal solution or minimizes the worst-case loss across tasks.

While effective, these gradient-level methods are fundamentally reactive; they resolve conflicts at each training step after they have already occurred. In contrast, our work offers a proactive, representation-centric solution. Rather than mediating conflicts during optimization, Domain Expansion constructs a latent space with inherent structural properties that prevent interference by design. Through its orthogonal pooling, our framework enforces that different objectives operate in separate, non-interfering subspaces. This proactive approach eliminates the need for runtime gradient manipulation and results in a latent space that is more explicit, interpretable, and robust to multi-objective training.

\begin{figure*}[t]
  \centering
  \includegraphics[width=0.95\textwidth,,height=0.12\textheight]{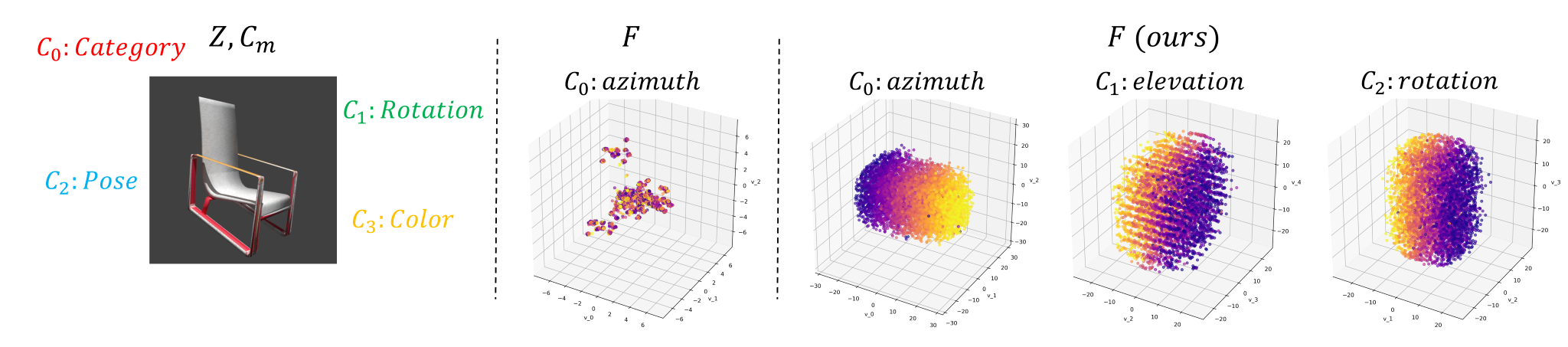}
    \caption{\textbf{Problem statement.} 
    (Left) Real-world inputs contain multiple concepts simultaneously. 
    (Center) Standard multi-objective training leads to \emph{latent representation collapse}, where concepts interfere and the latent space becomes entangled. 
    (Right) Our \emph{Domain Expansion} resolves this by assigning each concept to a mutually orthogonal subspace, yielding an explicit, interpretable, and compositional latent space.}
  \label{fig:problem}
\end{figure*}

\section{method}
Our goal is to design a latent space where a single representation can effectively encode information for multiple, competing objectives without interference. This section details our approach in three parts. First, we formalize the problem of latent representation collapse. Second, we introduce our solution, a framework we call Domain Expansion. Finally, we demonstrate how this method yields a structured and compositional latent space with powerful algebraic properties.

\subsection{Problem statement}
A standard representation learning pipeline involves an encoder, \(Enc(\cdot)\), that maps an input \(z\) from the data space \(\mathcal{Z}\) to a latent representation \(f \ \in \mathcal{F} \subset \mathbb{R}^D\). For each learning objective \(m\), a corresponding decoder, \(Dec(\cdot)\), then maps the latent representation \(f\) to a target concept \(\mathcal{C}\). The quality of this mapping is measured by an objective loss \(\mathcal{L}\). This can be expressed as:
\begin{equation}
\mathcal{Z} \xrightarrow{\text{Enc}} \mathcal{F} \xrightarrow{\text{Dec}} \mathcal{C} \quad (\text{evaluated by } \mathcal{L})
\label{eq:pipeline}
\end{equation}

In representation learning, the objective loss \(\mathcal{L}\) is often defined directly on the latent representations \(\mathcal{F}\). The loss function's goal is to impose a desired structure on the latent space (e.g., clustering or ranking), guided by the supervision from the target concept \(\mathcal{C}\).

The challenge arises when this framework is extended to handle a set of \(M\) objectives simultaneously. In this scenario, a single, shared representation \(\mathcal{F}\) is subjected to multiple objective losses, \(\{\mathcal{L}_m\}^{M-1}_{m=0}\), each guided by a distinct target concept, \(\{\mathcal{C}_m\}^{M-1}_{m=0}\). The total loss becomes a weighted sum applied to these shared representations:

\begin{equation}
\mathcal{L}_{\text{total}} = \sum_{m \in M} w_{m} \cdot \mathcal{L}_{m}(\mathcal{F}, \mathcal{C}_m).
\label{eq:total_collapse}
\end{equation}

As we have argued, naively minimizing this sum often leads to latent representation collapse, where conflicting structural demands from different losses force the shared representation \(\mathcal{F}\) into a compromised state that is suboptimal for all objectives (see Fig.~\ref{fig:problem}).

\begin{figure*}[t]
  \centering
  \includegraphics[width=0.7\textwidth,height=0.15\textheight]{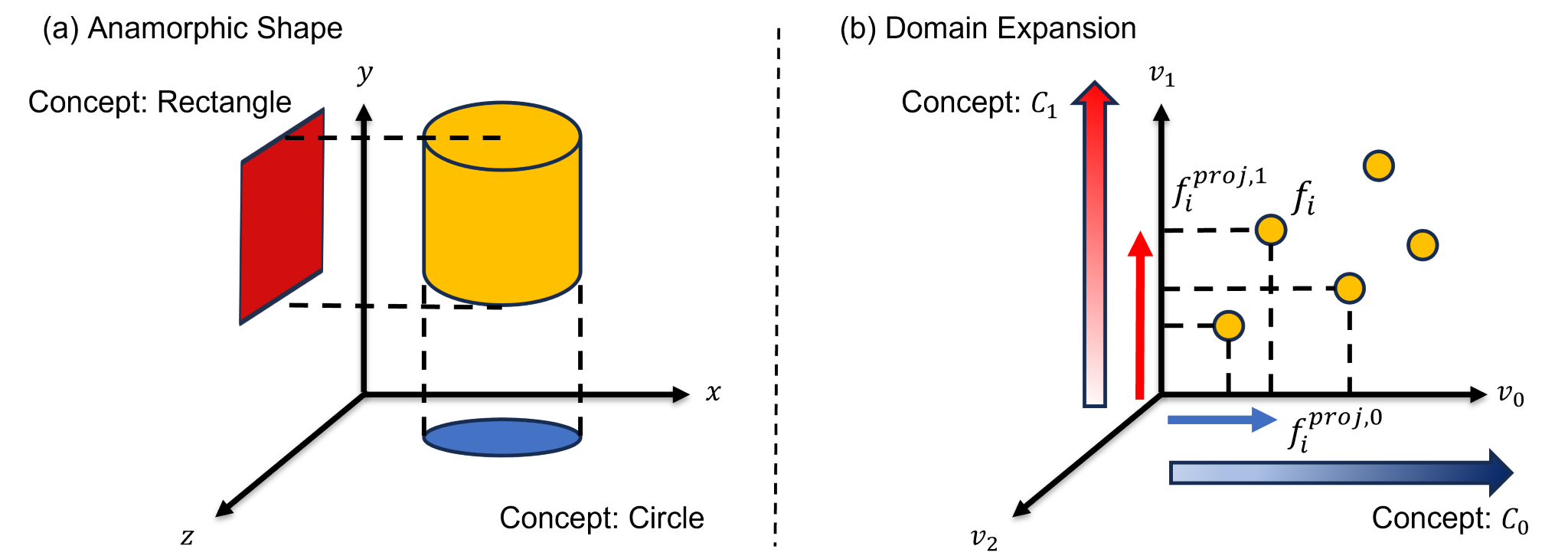}
    \caption{
A single latent vector represents multiple concepts through orthogonal projections, inspired by anamorphic art. (a) An anamorphic object, such as a cylinder, reveals different primitive concepts (a circle vs. a rectangle) when viewed from orthogonal directions. (b) Analogously, our method treats a single latent feature as a rich object that encodes multiple concepts simultaneously. The specific value for each concept is determined by its projection onto a corresponding orthogonal axis in the latent space.
}
  \label{idea}
\end{figure*}

\subsection{Domain Expansion}

To prevent latent collapse in multi-objective learning, we introduce Domain Expansion, a framework that structures the latent space $\mathcal{F}$ into a set of dedicated, orthogonal subspaces. The core idea is to ensure that the updates for each objective $m$ are confined to a unique subspace, preventing them from interfering with the representations of other objectives (see Fig. \ref{idea}). Our key insight is that the principal directions of variance in the latent space—its eigenvectors—can be harnessed to serve as an orthonormal basis for these dedicated subspaces.

The framework is a dynamic process applied at each training epoch, consisting of three steps:

\begin{enumerate}
    \item \textbf{Find Principal Axes.} First, we estimate the empirical mean $\mu$ and covariance $\Sigma$ of the latent feature distribution over the current batch or entire training set:
    \begin{equation}
        \mu = \mathbb{E}_{z\sim p(z)}[Enc(z)], \qquad \Sigma = \mathbb{E}_{z\sim p(z)}[(Enc(z)-\mu)(Enc(z)-\mu)^\top].
    \end{equation}
    We then perform an eigendecomposition of the covariance matrix to find an orthonormal basis of eigenvectors, $V = [v_0, v_1, \dots, v_{D-1}]$.
    \begin{equation}
        \Sigma = V\Lambda V^\top,\qquad V^\top V = I.
    \end{equation}
    \item \textbf{Define the Orthogonal Domain.} We select the top $M$ eigenvectors (those with the largest eigenvalues) to form our conceptual basis, which we call the \emph{domain}, $V_M = \{v_m \mid m \in M\}$. Each eigenvector $v_m$ is assigned then to represent a single target concept $\mathcal{C}_m$. This assignment ensures that the 1D subspace spanned by $v_m$ becomes exclusive for all projected features related to that concept. From this, we define the set of orthogonal subspaces and their corresponding projection operators:
    \begin{equation}
        \mathcal{F}^{\text{proj}}_m = \operatorname{span}(v_m), \quad \text{Proj}_m = v_m v_m^\top, \quad \forall m \in M.
    \end{equation}
    
    \item \textbf{Orthogonal Pooling.} After defining the orthogonal domain, we decompose the latent feature $f$ into each orthogonal, concept-specific subspace via computing its projection onto each of the domain's axes:
    \begin{equation}
        f^{\text{proj}, m} = \text{Proj}_m(f-\mu), \quad \forall m \in M.
    \end{equation}

    This step, which we term \emph{orthogonal pooling}, reformulates the learning pipeline as a one-to-many mapping from the shared space to the set of orthogonal subspaces, where each subspace corresponds to a single target concept.
    \begin{equation}
        \mathcal{Z} \xrightarrow{\text{Enc}} \mathcal{F} \xrightarrow{\text{Orthogonal Pooling}}
        \begin{cases}
            \mathcal{F}^{\text{proj}}_{0} \xrightarrow{\text{Dec}_0} \mathcal{C}_0 \\
            \mathcal{F}^{\text{proj}}_{1} \xrightarrow{\text{Dec}_1} \mathcal{C}_1 \\
            \qquad \vdots \\
            \mathcal{F}^{\text{proj}}_{M-1} \xrightarrow{\text{Dec}_{M-1}} \mathcal{C}_{M-1}
        \end{cases}
    \end{equation}
    The total training loss is then computed as the sum of individual losses on these independent, projected latent features:
    \begin{equation}
        \mathcal{L}_{\text{total}} = \sum_{m \in M} w_{m} \cdot \mathcal{L}_{m} (\mathcal{F}^{\text{proj}}_m, \mathcal{C}_m).
        \label{eq:de_total_loss}
    \end{equation}
\end{enumerate}
By decomposing the latent space into a set of dedicated, orthogonal subspaces for each concept, our framework ensures that the resulting loss gradients are inherently decoupled, preventing representation collapse by design.

\begin{figure*}[t]
  \centering
  \includegraphics[width=0.9\textwidth]{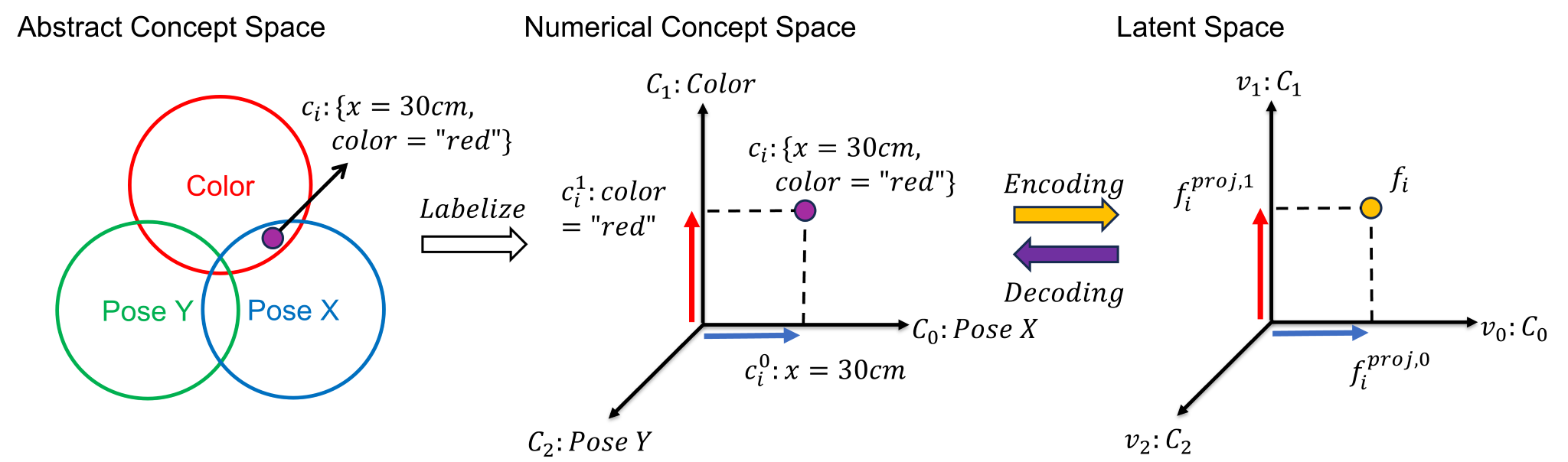}
    \caption{\textbf{Concept space vs. latent space:} (Left) Real-world inputs possess multiple abstract concepts simultaneously, such as color and pose. (Center) For training, we define a numerical concept space by assigning coordinates to these attributes. (Right) Our model then learns a mapping between this numerical space and its own internal, orthogonal latent space. This process creates an explicit structure where each axis corresponds to a single concept, allowing the model to robustly represent and manipulate the independent attributes of the input.
}
  \label{concept_n_latent_space}
\end{figure*}

\subsection{Properties and Operators of the Domain}

The orthogonal structure of our method is not merely a training aid; it endows the latent space with powerful properties, turning it into an interpretable and compositional \emph{concept algebra}.

To formalize this, we define our terminology in order:
\begin{enumerate}
    \item A \emph{target concept} $\mathcal{C}_m$ is the set of all possible values for a given attribute (e.g., Pose $X$).
    \item A \emph{target instantiated concept} $c^m$ is a single value from one such set, where $c^m \in \mathcal{C}_m$ (e.g., x = 30 cm).
    \item The \emph{concept space} $\mathcal{C}_{\text{space}}$ is the set of all possible combinations of attributes, formally defined as the Cartesian Product of the target concept sets (i.e. pose, color, category, etc.):
    \begin{equation}
    \mathcal{C}_{\text{space}} = \mathcal{C}_0 \times \mathcal{C}_1 \times \dots \times \mathcal{C}_{M-1} = \underset{m=0}{\overset{M-1}{\times}} \mathcal{C}_m.
    \end{equation}
    \item Finally, an \emph{instantiated concept} $c$ is a single element from this space, represented as a tuple $c = \{c^0, c^1, \dots, c^{M-1}\}$. During training, each input $z_i$ is associated with a specific instantiated concept $c_i$.
\end{enumerate}

For each target instantiated concept $c^m$ of an instantiated concept, we assume a mapping to a latent projection $f^{\text{proj}, m}$ via the decoder $Dec_m$. (The implementation of the decoder, $Dec_m$ and $Dec^{-1}_m$, is detailed in Appendix ~\ref{app:inverse_dec}).
\begin{equation}
c^m_i = \text{Dec}_m(f^{\text{proj}, m}_i), \quad f^{\text{proj}, m}_i = \text{Dec}^{-1}_m(c^m_i)
\label{eq:dec}
\end{equation}

\textbf{Property 1: Orthogonality of Target Concepts.}
Because each projection subspace $\mathcal{F}^{\text{proj}}_m$ is orthogonal to the others, the target concepts they represent are disentangled in the latent space:
\begin{equation}
\mathcal{F}^{\text{proj}}_0 \perp \mathcal{F}^{\text{proj}}_1 \perp \cdots \perp \mathcal{F}^{\text{proj}}_{M-1} \implies \mathcal{C}_0 \perp \mathcal{C}_1 \perp \cdots \perp \mathcal{C}_{M-1}.
\end{equation}

\textbf{Property 2: Multi-concept Encoding.}
A single latent feature $f_i$ simultaneously encodes a full instantiated concept $c_i$. The feature can be decomposed into its orthogonal projections, which are then decoded into their corresponding target instantiated concepts (see Fig. \ref{concept_n_latent_space}):
\begin{equation}
f_i \xrightarrow{\text{Pooling}} \{f^{\text{proj}, 0}_i, \dots, f^{\text{proj}, {M-1}}_i\} \xrightarrow{\text{Dec}} \{c^{0}_i, \dots, c^{M-1}_i\} \rightarrow c_i.
\end{equation}
Conversely, the full latent feature can be reconstructed from its components:
\begin{equation}
c_i \rightarrow \{c^{0}_i, \dots, c^{M-1}_i\} \xrightarrow{\text{Dec}^{-1}} \{f^{\text{proj}, 0}_i, \dots, f^{\text{proj}, {M-1}}_i\} \xrightarrow{\text{Reconst}} f_i,
\label{eq:reconstruction_flow}
\end{equation}
where the reconstruction is defined as:
\begin{equation}
f_i = \mu + \sum_{m\in M} f_i^{\text{proj}, m}.
\label{eq:reconstruction_formula}
\end{equation}
\textbf{Operator 1: Concept-Specific Adjustment ($\oplus^m$) and ($\ominus^m$).}
This operator adjusts an instantiated concept \(c_i\) by applying a change defined by a single target instantiated concept $c^m_\Delta \in \mathcal{C}_m$. The operation modifies the latent feature \(f_i\) without affecting any other target concepts. First, we find the latent vector for the change, $f^{\text{proj}, m}_{\Delta} = Dec^{-1}_m(c^m_{\Delta})$. The adjusted latent feature is then given by simple vector addition or subtraction:
\begin{equation}
f_j = f_i \pm f^{\text{proj}, m}_\Delta.
\end{equation}
The full derivation showing the correspondence between the concept and latent spaces is as follows:
\begin{align}
c_i \oplus^m c^m_\Delta & \rightarrow \{c^{0}_i, \dots, \{c^{m}_i \oplus^m c^{m}_{\Delta}\},\dots c^{M-1}_i\} \\
&\xrightarrow{\text{Dec}^{-1}} \{f^{\text{proj}, 0}_i, \dots, \{ f^{\text{proj}, m}_i + f^{\text{proj}, m}_\Delta\}, \dots, f^{\text{proj}, M-1}_i \} \\
&\xrightarrow{\text{Reconst}} f_i + f^{\text{proj}, m}_\Delta.
\end{align}
The derivation for the subtraction operator ($\ominus^{m}$) is analogous.

\textbf{Operator 2: Concept Composition ($\oplus$) and ($\ominus$).}
This operator composes two full instantiated concepts, $c_p$ and $c_q$, by operating on their corresponding latent representations, \(f_p\) and \(f_q\). The composition is achieved through simple vector addition or subtraction:
\begin{equation}
f_{pq} = f_p \pm f_q.
\end{equation}
This operation corresponds to a component-wise combination in each orthogonal subspace:
\begin{align}
c_{p} \oplus c_{q} &\rightarrow \{\{c^{0}_p \oplus^0 c^{0}_q \}, \{c^{1}_p \oplus^1 c^{1}_q \}, \dots, \{c^{M-1}_p \oplus^{M-1} c^{M-1}_q \}\} \\
&\xrightarrow{Dec^{-1}} \{\{f^{proj, 0}_p + f^{proj, 0}_q \}, \{f^{proj, 1}_p + f^{proj, 1}_q \}, \dots, \{f^{proj, M-1}_p + f^{proj, M-1}_q \}\} \\
&\xrightarrow[]{\text{Reconst}} f_p +f_q.
\end{align}
The derivation for the subtraction operator ($\ominus$) is analogous.

\section{Experiments}

To validate our framework, we formulated and tested three key hypotheses (H):


\begin{enumerate}
    \item[{H1:}] Does training a single network with multiple objectives lead to \textbf{latent representation collapse} as expected? 
    \item[{H2:}] Does \textbf{Domain Expansion} prevent this collapse and outperform standard multi-task learning baselines? 
    \item[{H3:}] Does our method create a truly \textbf{compositional and inferable} latent space, as claimed?
\end{enumerate}

Through designing experiments to test these hypotheses, we aimed to systematically evaluate our framework's effectiveness and validate its core assumptions. Specifically, we conduct experiments on the ShapeNet dataset \citep{shapenet}, a standard benchmark for 3D object classification and pose estimation. We further validate our method on the MPIIGaze \citep{MPIIGaze} and Rotated MNIST \citep{MNIST} datasets (see Appendix \ref{app:experiments} for full details).

\subsection{Experimental Setup}

\begin{figure*}[t]
  \centering
  \includegraphics[width=0.95\textwidth]{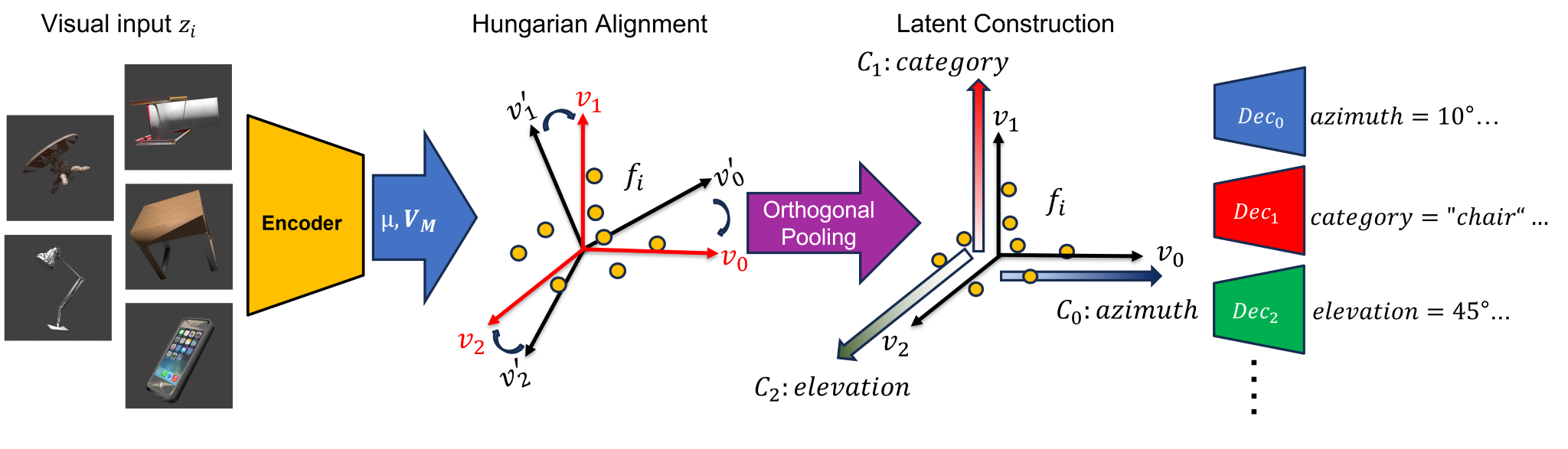}
    \caption{\textbf{System overview:} An input image is first passed through an encoder to produce a latent feature. From the distribution of these features across the dataset, we compute the mean ($\mu$) and an orthogonal basis of eigenvectors ($V_M$). To ensure consistency during training, this basis is stabilized across epochs using Hungarian alignment. The orthogonal pooling then projects each latent feature onto these basis vectors. Finally, these projected, non-interfering representations are fed into concept-specific decoders to produce the final outputs.
}
  \label{overview}
\end{figure*}

\textbf{Architecture and Training.} We adopt an encoder-decoder architecture here. The encoder, $Enc(\cdot)$, is a ResNet-50 \citep{ResNet} backbone that produces a 2048-dimensional latent feature, $f \in \mathcal{F}$. For each of the $M$ objectives, its corresponding decoder, $Dec_m(\cdot)$, is a single linear layer. Training proceeds in two stages: first, we train the encoder while dynamically updating the orthogonal basis ($V_M$ and $\mu$) at each epoch. To stabilize the basis during early training, we align the eigenvectors between epochs using the Hungarian algorithm \citep{hungarian} (see Appendix \ref{app:hungarian}). Second, once the latent space has converged, we freeze the encoder and train the simple linear decoders on the final, fixed representations (see Fig.~\ref{overview}).

\textbf{Dataset and Concepts.} We use the ShapeNet dataset, rendering approximately 30,000 images from 10 models per category using the 3D-R2N2 pipeline \citep{3D-R2N2}. We define five target concepts: three for regression—azimuth (sampled from $[-\pi/2, \pi/2]$), elevation, and in-plane rotation (both sampled from $[-\pi/4, \pi/4]$)—and two for classification (object category and model ID). We denote these concepts as $\{\mathcal{C}_{\text{az}}, \mathcal{C}_{\text{el}}, \mathcal{C}_{\text{rot}}, \mathcal{C}_{\text{cat}}, \mathcal{C}_{\text{id}}\}$. 

\textbf{Loss Functions.} To structure the latent space, we use a weighted combination of two representation learning losses that operate on the high-dimensional latent vectors. Supervised Contrastive (SupCon) \citep{SupCon} imposes a clustering constraint using a binary mechanism (positive vs. negative pairs). We adapt the standard SupCon loss by replacing the inner product similarity with L2 distance, which is more compatible with our projection-based method. Rank-N-Contrast (RNC) \citep{RNC} enforces a more fine-grained ranking mechanism. For both losses, we set the temperature $\tau = 2.0$. The final loss weights $w_m$ are to 1.0 for RNC and 0.02 for SupCon.

Based on these components, we define two primary sets of objectives for our experiments, which correspond in order to the concepts $\{\mathcal{C}_{\text{az}}, \mathcal{C}_{\text{el}}, \mathcal{C}_{\text{rot}}, \mathcal{C}_{\text{cat}}, \mathcal{C}_{\text{id}}\}$. \textbf{Objective Set 1} uses the RNC loss for all five concepts. \textbf{Objective Set 2} uses RNC for the first three concepts (regression) and SupCon for the final two (classification).


\subsection{Evaluating Multi-Objective Performance}
\label{sec:performance_eval}

To answer our first two research questions (H1 and H2), we evaluate our method against a suite of baselines. We establish a simple baseline trained with a weighted sum of all losses (Eq.~\ref{eq:total_collapse}). We also compare against three gradient-based MTL methods designed to mitigate gradient conflict: Nash-MTL \citep{Nash-MTL}, FAMO \citep{FAMO}, and IMTL \citep{IMTL}. Finally, we evaluate the performance of our full proposed method, Domain Expansion (Ours).

We evaluate the quality of the learned representations for each concept using metrics tailored to the task type. For regression concepts (azimuth, elevation, rotation), we measure the Spearman's rank correlation \citep{spearman}. For classification concepts (category, ID), we measure the V-measure (V-score) \citep{v-measure} to evaluate cluster quality. We also report standard predictive metrics: Mean Absolute Error (MAE$^{\circ}$) for regression and accuracy (Acc.) for classification.

The results in Table~\ref{tab:shapenet_comprehensive_metrics} demonstrate the effectiveness of our Domain Expansion framework in preventing latent representation collapse. The baseline model confirms that naive multi-objective training leads to a disorganized latent space, exhibiting poor performance on representation quality metrics like Spearman correlation and V-score. A particularly revealing result is visible for Objective Set 2, where several baselines achieve high classification accuracy but have a V-score near zero. This discrepancy highlights that they learn a superficial shortcut for the predictive task while their latent space remains collapsed.

\begin{figure*}[t]
  \centering
  \includegraphics[width=0.9\textwidth]{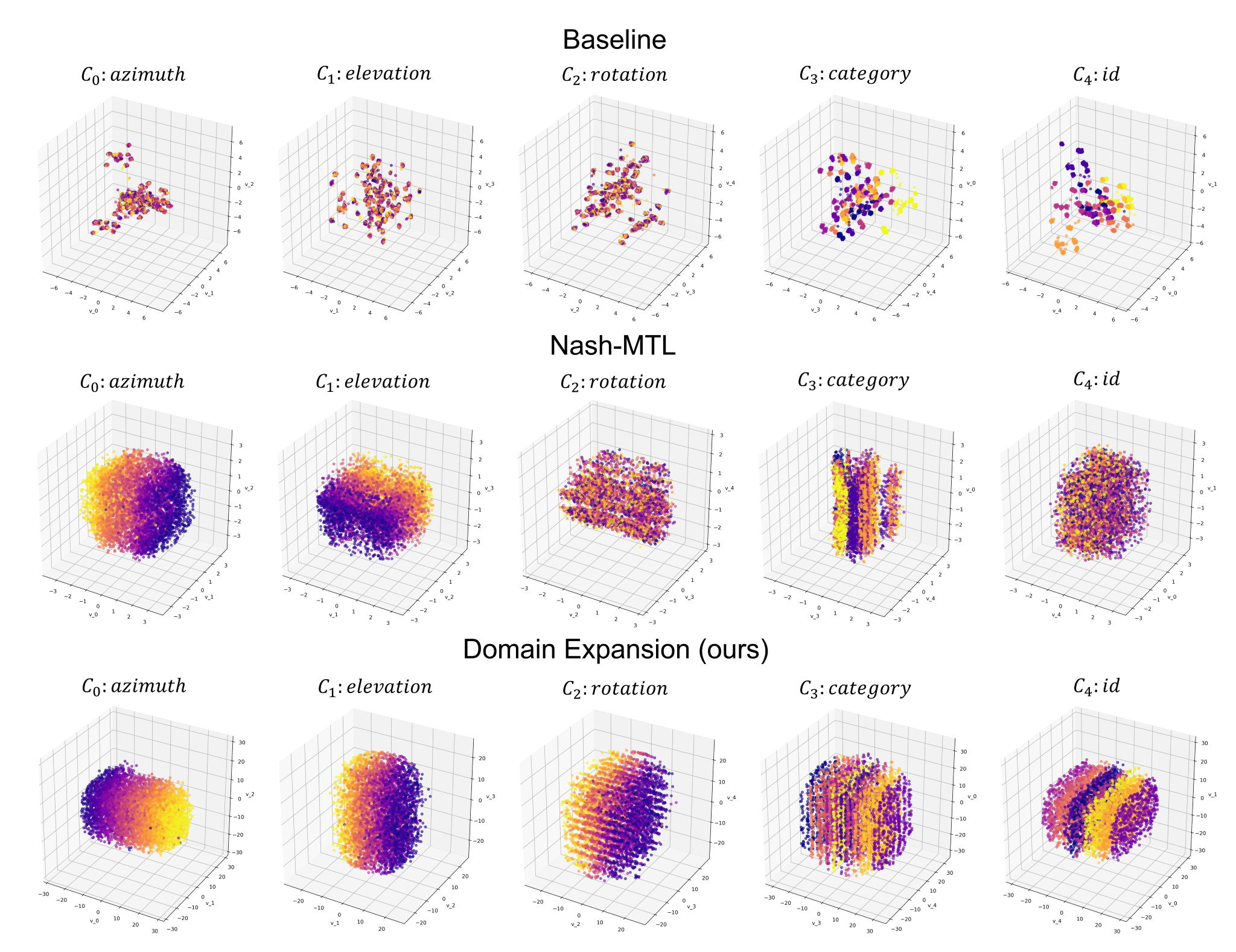}
    \caption{\textbf{Qualitative Comparison of Latent Space Structures.} We visualize the latent spaces from the baseline, Nash-MTL, and Ours (top to bottom rows) via PCA. Points are colored by their ground-truth concept value, from low (dark purple) to high (bright yellow). Our method (bottom row) demonstrates a distinctly more organized structure with clear directional alignment for each concept, unlike the entangled representations produced by the baseline methods.}
  \label{fig:latent_space}
\end{figure*}

In contrast, Domain Expansion significantly outperforms all baselines, achieving superior scores on both the representation metrics and the final predictive tasks. This quantitative success is mirrored in our qualitative visualizations in Fig.~\ref{fig:latent_space}. Whereas the latent spaces of baseline methods appear entangled and unstructured, the space learned by our method is clearly organized, with concepts aligning along their corresponding orthogonal axes.


\subsection{Probing the Compositional Latent Space}
\label{sec:compositional_eval}

Finally, we designed an experiment to test the \textbf{composition operator} ($\oplus$ and $\ominus$) and verify that our latent space is truly inferable (H3). The goal is to see if we can transform one latent vector $f_p$ into a target vector $f_q$ by applying a conceptual difference defined in the concept space.

\begin{enumerate}
    \item We split the test set into two halves, P and Q. For a pair of samples $(z_p, c_p)$ and $(z_q, c_q)$, our goal is to synthetically reconstruct $f_q$.
    \item The ground-truth target vector is obtained directly from the encoder: $f_q = \text{Enc}(z_q)$.
    \item We define the conceptual difference between the two samples as $c_\Delta = c_q \ominus c_p$.
    \item We then create a synthetic target concept by applying this difference to our source concept: $c_q^* = c_p \oplus c_\Delta$.
    \item Using our framework's operators, we reconstruct the latent vector corresponding to this synthetic concept: $f_q^* = \text{Reconst}(c_q^*)$ (Eq.~\ref{eq:reconstruction_flow}).
    \item We evaluate the quality of this synthetic reconstruction by computing the average cosine similarity between the reconstructed vectors and the ground-truth vectors over all pairs in the test set Q:
    \(\mathbb{E}_{(z_p, z_q)}[\cos(f_q, f_q^*)]\).
\end{enumerate}
As shown in the final column of Table~\ref{tab:shapenet_comprehensive_metrics}, our method's concept composition performance, measured by cosine similarity, is substantially higher than all baselines. This result confirms that our latent space is not a ``black box," but rather a meaningful, compositional structure where conceptual operations correspond to simple vector arithmetic, enabling accurate inference and manipulation.


\begin{table}[t]
\centering
\small
\caption{Comprehensive comparison of representation quality, predictive performance, and concept composition. Arrows indicate whether higher ($\uparrow$) or lower ($\downarrow$) values are better.}
\label{tab:shapenet_comprehensive_metrics}
\footnotesize 
\setlength{\tabcolsep}{3.2pt}
\begin{tabular}{ll ccc cc ccc cc c}
\toprule
\multirow{3}{*}{Objective Set} & \multirow{3}{*}{Method} & \multicolumn{10}{c}{Representation \& Predictive Performance} & \multicolumn{1}{c}{Concept Comp.} \\
\cmidrule(lr){3-12} \cmidrule(lr){13-13}
& & \multicolumn{3}{c}{Spearman $\uparrow$} & \multicolumn{2}{c}{V-score $\uparrow$} & \multicolumn{3}{c}{MAE$^{\circ}$ $\downarrow$} & \multicolumn{2}{c}{Acc. $\uparrow$} & \multicolumn{1}{c}{Sim. $\uparrow$} \\
\cmidrule(lr){3-5}\cmidrule(lr){6-7}\cmidrule(lr){8-10}\cmidrule(lr){11-12}\cmidrule(lr){13-13}
& & az & el & rot & cat & id & az & el & rot & cat & id & $\oplus$ and $\ominus$\\
\midrule
\multirow{5}{*}{Objective Set 1}
& baseline  & 0.41 & 0.34 & 0.35 & 0.16   & 0.14   & 0.12 & 0.09 & 0.09 & 0.28 & 0.37 & 0.22   \\
& FAMO      & 0.49 & 0.41 & 0.42 & 0.00 & 0.00 & 0.12 & 0.09 & 0.09 & 0.19 & 0.18 & 0.28   \\
& Nash-MTL      & 0.38 & 0.41 & 0.42 & 0.00 & 0.00 & 0.11 & 0.09 & 0.09 & 0.17 & 0.13 & 0.28   \\
& IMTL      & 0.31 & 0.16 & 0.16 & 0.39 & 0.28 & 0.14 & 0.11 & 0.12 & 0.92 & 0.79 & 0.14   \\
& \textbf{Ours} & 0.95 & 0.87 & 0.85 & 0.99 & 0.91 & 0.08 & 0.08 & 0.09 & 0.99 & 0.97 & 0.95 \\
\midrule
\multirow{5}{*}{Objective Set 2}
& baseline  & 0.01 & 0.01 & 0.01 & 0.99   & 0.00   & 0.77 & 0.38 & 0.38 & 0.99 & 0.99 & 0.42   \\
& FAMO      & 0.28 & 0.23 & 0.22 & 0.99 & 0.00 & 0.19 & 0.14 & 0.13 & 0.99 & 0.99 & 0.28   \\
& Nash-MTL      & 0.45 & 0.39 & 0.39 & 0.15 & 0.00 & 0.12 & 0.08 & 0.09 & 0.99 & 0.99 & 0.35   \\
& IMTL      & 0.39 & 0.18 & 0.16 & 0.99 & 0.00 & 0.15 & 0.11 & 0.13 & 0.99 & 0.99 & 0.28   \\
& \textbf{Ours} & 0.95 & 0.87 & 0.85 & 0.98 & 0.96 & 0.07 & 0.08 & 0.09 & 0.98 & 0.94 & 0.93 \\
\bottomrule
\end{tabular}
\end{table}

\section{Discussion and limitations}
A key design choice in our framework is leveraging projection-based supervision. Unlike other approaches that might constrain the entire latent vector, our method only guides its components along the pre-defined orthogonal axes. It is innately permissive by design: it provides the encoder with significantly more degrees of freedom to learn a rich internal representation, which we believe contributes to the performance and compositional properties we observed.

The primary strength of such a structured space is its support for high-level conceptual compositions that are difficult for human beings to understand intuitively. For instance, our framework can systematically represent a novel concept like ``chair" $\oplus$ ``boat" via the simple latent space operation \(f_{chair} + f_{boat}\). Our current limitation, therefore, is not in representation but in decoding these abstract concepts. A promising direction for future work is to pair our encoder with a generative model, such as an LLM or diffusion model, to interpret these latent compositions into human-understandable outputs.

\section{Conclusion}
In this work, we introduced Domain Expansion, a framework that addresses latent representation collapse by constructing a latent space with dedicated orthogonal subspaces for each task. Our experiments demonstrated that this design alleviates the risk of task interference and yields an explicit and compositional latent space that supports algebraic concept manipulation. This approach suggests a future pathway for a more structured bridge between high-level concepts and a model's learned representation, laying a promising foundation for more
controllable and interpretable models with applications in areas like algorithmic fairness and controllable multi-modal content generation.

\section{Acknowledgment}
\label{sec:Acknowledgment}

This research is sponsored by NSF, the Partnerships for Innovation grant (\#2329780) and TRINA Mothership project. We thank the Research Computing (RC) at Arizona State University (ASU) \citep{ASUSOL} and the NSF NAIRR program for their generous support in providing
computing resources. The views and opinions of the authors
expressed herein do not necessarily state or reflect those of
the funding agencies and employers.

\bibliography{iclr2026_conference}
\bibliographystyle{iclr2026_conference}

\clearpage
\appendix
\section{Appendix}

\subsection{On the Invertibility of the Decoder}
\label{app:inverse_dec}

In our method, the concept operators require a mapping from a concept instance back to its latent representation, denoted as $f^{\text{proj}, m} = \text{Dec}_m^{-1}(c^m)$. This section details how we implement this inverse mapping.

Our decoders are single linear layers:
\begin{equation}
c^m = W \cdot f^{\text{proj}, m} + b,
\end{equation}
where $W$ and $b$ are the layer's weights and bias. Since the latent space dimension ($D=2048$) is much larger than the concept dimension (e.g., 1 for a regression target or the number of classes (logit) for a classification target), the weight matrix $W$ is non-square and a standard inverse does not exist. The inverse problem is \textbf{ill-posed}: for any given output $c^m$, there is an entire affine subspace of possible inputs that could have produced it.

However, our Domain Expansion framework provides a powerful geometric constraint that makes this problem tractable. By construction, we know that any valid projected feature $f^{\text{proj}, m}$ must lie on the 1D subspace spanned by its corresponding eigenvector, $v_m$. This constraint resolves the ambiguity of the inverse mapping.

\begin{figure}[h]
\centering
\includegraphics[width=0.4\textwidth]{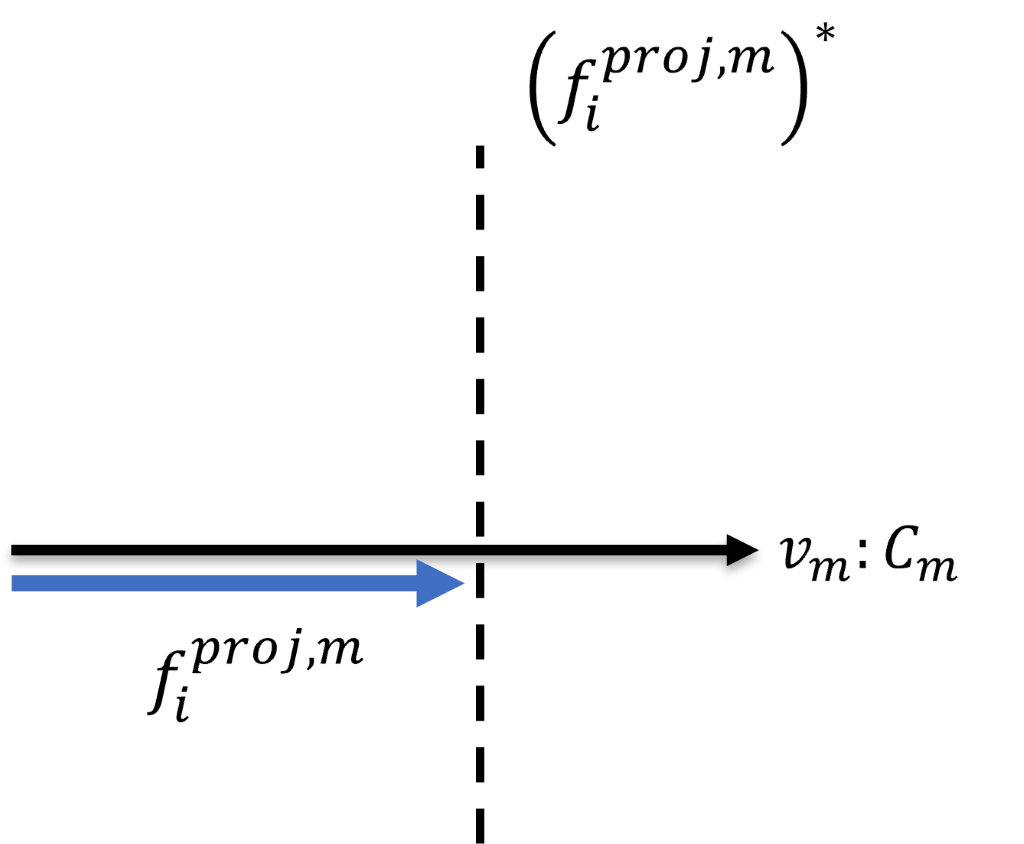}
\caption{\textbf{Recovering a unique inverse for the decoder.} The set of all possible latent features that could produce a given output forms a high-dimensional preimage (illustrated by the dashed line). Our method provides a strong constraint: the true solution must lie on the subspace spanned by the eigenvector $v_m$. By projecting an arbitrary solution $(f^{\text{proj}, m})^*$ onto this subspace, we can recover the unique, correct latent feature $f^{\text{proj}, m}$.}
\label{fig:inverse_dec}
\end{figure}

Therefore, we can implement the inverse decoder as a two-step process, as illustrated in Figure~\ref{fig:inverse_dec}. First, we find an arbitrary solution, $(f^{\text{proj}, m})^*$, in the high-dimensional preimage (e.g., using the pseudo-inverse). Second, we simply project this arbitrary solution back onto the correct eigenvector to recover the unique, valid latent feature:
\begin{equation}
f^{\text{proj}, m} = \text{Proj}_m((f^{\text{proj}, m})^*).
\end{equation}
This elegant solution is a direct benefit of the explicit structure imposed by our method.

\subsection{Subspace Stability and Alignment}
\label{app:hungarian}

During iterative training, the eigenvectors that define our conceptual subspaces can be subject to two forms of ambiguity. First, the order of the learned eigenvectors may permute between epochs. Second, an eigenvector $v$ and its negative $-v$ span the same 1D subspace. These ambiguities can destabilize training as the network may associate a concept with a different eigenvector at each step.

To resolve this, we enforce consistency across training epochs using the Hungarian algorithm. At the end of each epoch, we compute a pairwise cosine similarity matrix between the newly learned eigenvectors and the eigenvectors from the previous epoch. The Hungarian algorithm then finds the optimal one-to-one assignment that maximizes the similarity, resolving any permutation ambiguity. We also align the sign of each eigenvector to ensure the cosine similarity is positive.

\begin{figure}[htbp]
\centering
\includegraphics[width=0.95\textwidth]{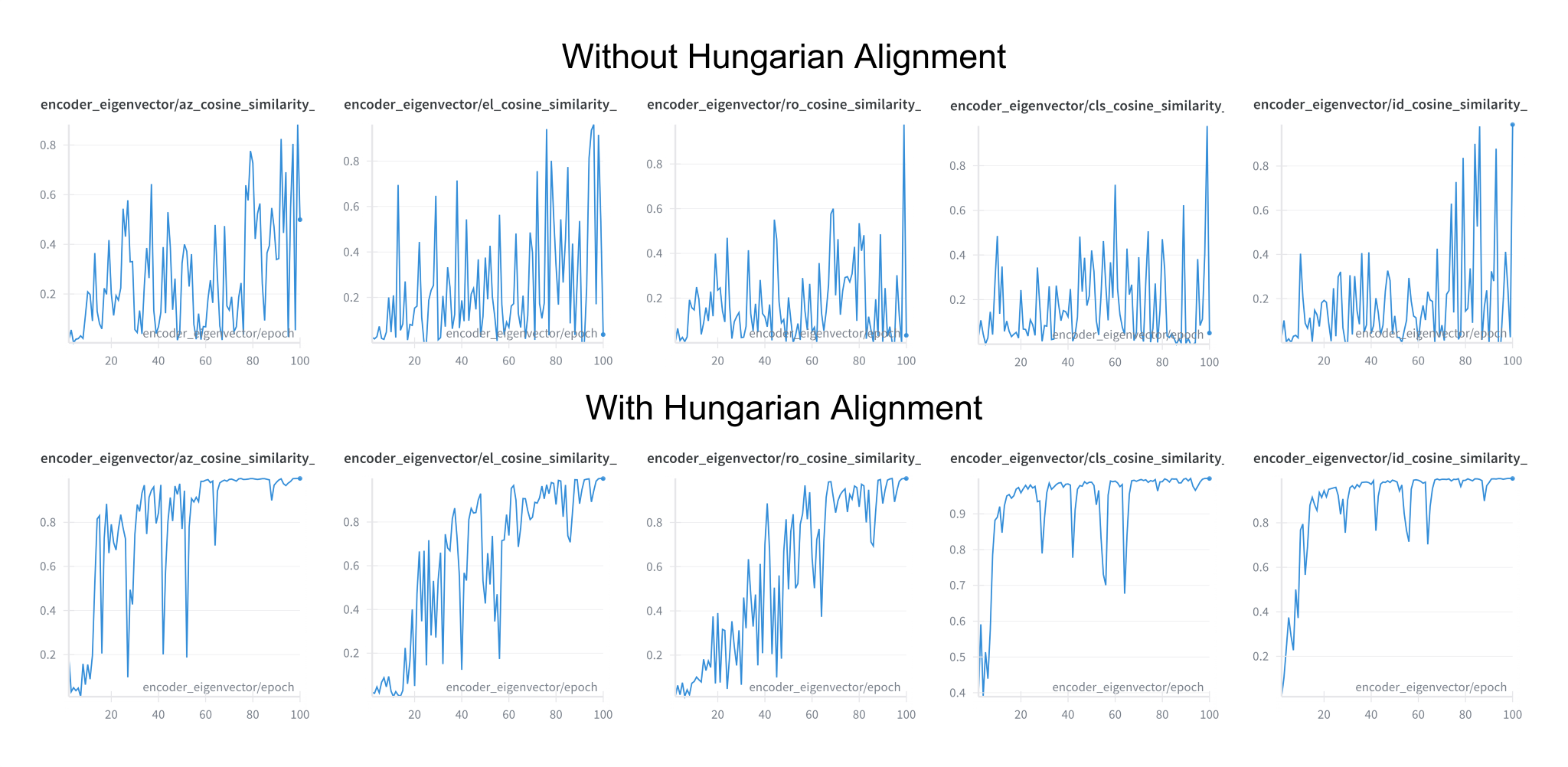}
\caption{\textbf{Eigenvector stabilization during training.} We plot the cosine similarity between the aligned eigenvectors for each concept from the current and previous training epochs. The increasing similarity demonstrates the convergence of the learned subspaces. From left to right, the plots correspond to the subspaces for azimuth, elevation, in-plane rotation, category, and ID, respectively.}
\label{fig:hungarian} 
\end{figure}

This alignment procedure ensures that the learned subspaces stabilize and converge. Figure~\ref{fig:hungarian} visualizes this process, showing the cosine similarity between aligned eigenvectors from consecutive epochs. The similarities approach 1.0, indicating convergence. Table~\ref{tab:hungarian} confirms this, showing that the final similarities are nearly perfect, which demonstrates that the conceptual subspaces are stable upon the completion of training.

\begin{table}[htbp]
\centering
\small
\caption{Effect of Hungarian Alignment on Eigenvector Stability}
\label{tab:hungarian_ablation}
\begin{tabular}{l|ccccc}
\toprule
Method & Azim. & Elev. & Rot. & Cat. & ID \\
\midrule
Without Hungarian & 0.49 & 0.04 & 0.03 & 0.05 & 0.98 \\
With Hungarian & 0.99 & 0.99 & 0.99 & 0.99 & 0.99 \\
\bottomrule
\end{tabular}
\label{tab:hungarian}
\end{table}

\subsection{Experiment result of other dataset}
\label{app:experiments}

To demonstrate the generalizability of our Domain Expansion framework, we conduct experiments on two additional, diverse benchmarks: the MPIIGaze dataset for fine-grained gaze estimation and a custom Rotated MNIST dataset for classification under pose variation.

\textbf{MPIIGaze Setup.} The MPIIGaze dataset \citep{MPIIGaze} contains approximately 30,000 images from 15 participants for the task of gaze estimation. Following our setup for ShapeNet, we use a ResNet-50 backbone. We define four target concepts corresponding to the ordered set  $\{\mathcal{C}_{\text{x}}, \mathcal{C}_{\text{y}}, \mathcal{C}_{\text{z}}, \mathcal{C}_{id}\}$: the three axes of 3D gaze direction (x,y,z) for regression, and the participant ID for classification.

Based on this ordering, we evaluate two objective sets. Objective Set 1 uses the RNC loss for all four concepts. Objective Set 2 uses RNC for the three regression concepts and SupCon for the classification concept. The complete results for these experiments, detailing representation quality, predictive performance, and concept composition, are summarized in Table~\ref{tab:mpiigaze_comprehensive_metrics}.

\textbf{Rotated MNIST Setup.} To further test our model's ability to handle pose concepts, we created a Rotated MNIST dataset by applying random rotations in the range $[-\pi/2, \pi/2]$ to images from the original MNIST dataset \citep{MNIST}. For this task, we use a ResNet-18 backbone and define two concepts corresponding to the ordered set $\{\mathcal{C}_{rot}, \mathcal{C}_{id}\}$: the rotation angle for regression and the digit identity for classification.

Based on this, we evaluate two objective sets. Objective Set 1 uses the RNC loss for both concepts. Objective Set 2 uses RNC for the regression concept and SupCon for the classification concept. The complete results are presented in Table~\ref{tab:mnist_comprehensive_metrics}.

\begin{table}[htbp]
\centering
\caption{Comprehensive comparison of representation quality, predictive performance, and concept composition on the MPIIGaze dataset.}
\label{tab:mpiigaze_comprehensive_metrics}
\footnotesize 
\setlength{\tabcolsep}{3.8pt}
\begin{tabular}{ll ccc c ccc c c}
\toprule
\multirow{3}{*}{Objective Set} & \multirow{3}{*}{Method} & \multicolumn{8}{c}{\textbf{Representation \& Predictive Performance}} & \multicolumn{1}{c}{\textbf{Concept Comp.}} \\
\cmidrule(lr){3-10} \cmidrule(lr){11-11}
& & \multicolumn{3}{c}{Spearman $\uparrow$} & \multicolumn{1}{c}{V-score $\uparrow$} & \multicolumn{3}{c}{MAE $\downarrow$} & \multicolumn{1}{c}{Acc. $\uparrow$} & \multicolumn{1}{c}{Sim. $\uparrow$} \\
\cmidrule(lr){3-5}\cmidrule(lr){6-6}\cmidrule(lr){7-9}\cmidrule(lr){10-10}\cmidrule(lr){11-11}
& & x & y & z & id & x & y & z & id & $\oplus$ and $\ominus$ \\
\midrule
\multirow{5}{*}{Objective Set 1}
& baseline  & 0.54 & 0.48 & 0.53 & 0.18 & 0.01 & 0.02 & 0.02 & 0.49 & 0.14   \\
& FAMO      & 0.52   & 0.40   & 0.52   & 0.23   & 0.01   & 0.03   & 0.03   & 0.18   & 0.09   \\
& Nash-MTL   & 0.49   & 0.40   & 0.51   & 0.25  & 0.01   & 0.02   & 0.03  & 0.36 & 0.09   \\
& IMTL      & 0.15   & 0.44   & 0.20   & 0.39   & 0.02   & 0.03   & 0.03   & 0.85   & 0.07   \\
& \textbf{Ours} & 0.72 & 0.92 & 0.79 & 0.99 & 0.01 & 0.02 & 0.02 & 0.99 & 0.95 \\
\midrule
\multirow{5}{*}{Objective Set 2}
& baseline  & 0.29 & 0.16 & 0.21 & 0.99 & 0.01 & 0.04 & 0.06 & 0.99 & 0.63   \\
& FAMO      & 0.50   & 0.37   & 0.47   & 0.99   & 0.01   & 0.03   & 0.03   & 0.99   & 0.36   \\
& Nash-MTL      & 0.36   & 0.45   & 0.32   & 0.99   & 0.01   & 0.03   & 0.03   & 0.99   & 0.29   \\
& IMTL      & 0.28   & 0.51   & 0.29   & 0.99   & 0.01   & 0.03   & 0.03   & 0.99   & 0.35   \\
& \textbf{Ours} & 0.73 & 0.92 & 0.81 & 0.98 & 0.01 & 0.02 & 0.01 & 0.98 & 0.89 \\
\bottomrule
\end{tabular}
\end{table}

\begin{table}[htbp]
\centering
\caption{Comprehensive comparison of representation quality, predictive performance, and concept composition on the Rotated MNIST dataset.}
\label{tab:mnist_comprehensive_metrics}
\footnotesize 
\setlength{\tabcolsep}{4pt}
\begin{tabular}{ll c c c c c}
\toprule
\multirow{3}{*}{Objective Set} & \multirow{3}{*}{Method} & \multicolumn{4}{c}{\textbf{Representation \& Predictive Performance}} & \multicolumn{1}{c}{\textbf{Concept Comp.}} \\
\cmidrule(lr){3-6} \cmidrule(lr){7-7}
& & \multicolumn{1}{c}{Spearman $\uparrow$} & \multicolumn{1}{c}{V-score $\uparrow$} & \multicolumn{1}{c}{MAE (rad) $\downarrow$} & \multicolumn{1}{c}{Acc. $\uparrow$} & \multicolumn{1}{c}{Sim. $\uparrow$} \\
\cmidrule(lr){3-3}\cmidrule(lr){4-4}\cmidrule(lr){5-5}\cmidrule(lr){6-6}\cmidrule(lr){7-7}
& & rot & id & rot & id & $\oplus$ and $\ominus$ \\
\midrule
\multirow{5}{*}{Objective Set 1}
& baseline  & 0.63 & 0.47 & 0.14 & 0.78 & 0.39  \\
& FAMO      & 0.67   & 0.44  & 0.13  & 0.84   & 0.37   \\
& Nash-MTL      & 0.62  & 0.42 & 0.14 & 0.82 & 0.38   \\
& IMTL      & 0.57  & 0.47 & 0.14  & 0.87  & 0.39   \\
& \textbf{Ours} & 0.92 & 0.87 & 0.14 & 0.88 & 0.73 \\
\midrule
\multirow{5}{*}{Objective Set 2}
& baseline  & 0.02 & 0.96 & 0.73 & 0.99 & 0.71   \\
& FAMO      & 0.59   & 0.96  & 0.14  & 0.99   & 0.54 \\
& Nash-MTL      & 0.02  & 0.96  & 0.74 & 0.99  & 0.45   \\
& IMTL      & 0.71  & 0.95   & 0.14 & 0.98 & 0.48   \\
& \textbf{Ours} & 0.93 & 0.96 & 0.12 & 0.99 & 0.71 \\
\bottomrule
\end{tabular}
\end{table}

\subsection{Continual Learning with Domain Expansion}
\label{sec:appendix_continual}

Our main paper assumes a fixed set of $M$ tasks defined \textit{a priori}. A natural question is whether Domain Expansion can adapt to a continual learning setting where new tasks are added to a pre-trained model. We designed an experiment to validate this, demonstrating that our framework can successfully add $N$ new objectives ($C_{M+1}, ..., C_{M+N}$) to a model already pre-trained on $M$ tasks, without retraining from scratch.

Our continual learning approach is as follows:
\begin{enumerate}
    \item \textbf{Freeze Existing Axes:} We "freeze" the original $M$ axes ($v_0...v_{M-1}$), which have already been trained.
    
    \item \textbf{Find $N$ New Orthogonal Axes:} We find $N$ new axes by running eigendecomposition on the "residual" feature space $\mathcal{F}'$. This residual space is computed by removing the components from the $M$ pre-trained axes: 
    $$ f' = f - \sum_{i=0}^{M-1} f^{\text{proj}, i} $$
    We select the top $N$ eigenvectors from the covariance of $\mathcal{F}'$, which guarantees they are orthogonal to all previously learned axes.
    
    \item \textbf{Train New Tasks (with Regularization):} We then train the $N$ new tasks. Each new objective $C_{j}$ (where $j \in \{M+1, ..., M+N\}$) is trained using its loss $\mathcal{L}_{j}$ on projections onto its corresponding new axis $v_{j}$.
    
    \item \textbf{Prevent Catastrophic Forgetting:} To prevent the encoder from "forgetting" the original tasks, we add an L2 regularization loss. This loss ensures that the projected coefficients for the original $M$ tasks (on their frozen axes $v_0...v_{M-1}$) remain constant.
\end{enumerate}

The total loss function for this new training phase, where $\mathcal{N}$ is the set of $N$ new tasks and $\mathcal{M}$ is the set of $M$ old tasks, is:

\begin{equation}
    \mathcal{L}_{\text{total}} = \underbrace{\sum_{j \in \mathcal{N}} w_{j} \cdot \mathcal{L}_{j} (\mathcal{F}^{\text{proj}}_j, \mathcal{C}_j)}_{\text{Loss for N new tasks}} + \underbrace{\lambda \sum_{m \in \mathcal{M}} \mathcal{L}_{\text{L2}} (\mathcal{E}_{\text{coeffs}, m}, \bar{\mathcal{E}}_{\text{coeffs}, m})}_{\text{Regularization for M old tasks}}.
    \label{eq:de_continual_loss}
\end{equation}

Here, $\mathcal{L}_{\text{L2}}$ is the L2 (Mean Squared Error) loss, $\lambda$ is a balancing hyperparameter ($\lambda$ = 0.1), $\mathcal{E}_{\text{coeffs}, m}$ is the vector of coefficients for the $m$-th original task from the current encoder, and $\bar{\mathcal{E}}_{\text{coeffs}, m}$ is the "frozen" target vector of coefficients captured before this new training began.

The results of this experiment are presented in Table~\ref{tab:shapenet_continual_learning}. The "Continual" model, which was fine-tuned, shows a minor performance drop compared to the "Default" model trained on all tasks from scratch. This drop also affects the concept composition quality, which decreases from 0.95 to 0.72. However, the framework still successfully solves the latent collapse problem and prevents catastrophic forgetting. Crucially, even with this degradation, the model's performance on both representation (e.g., Spearman, V-score) and compositionality (0.72) remains remarkably stronger than the baselines in Table 1 (main paper), demonstrating the framework's viability for continual learning.

\begin{table}[htbp]
\centering
\small
\caption{Performance comparison of models trained from scratch versus continually. The "Continual" model was pre-trained on $\{az, el, cat\}$, then fine-tuned on $\{rot, id\}$. Arrows indicate whether higher ($\uparrow$) or lower ($\downarrow$) values are better.}
\label{tab:shapenet_continual_learning}
\footnotesize 
\setlength{\tabcolsep}{3.2pt}
\begin{tabular}{ll ccc cc ccc cc c}
\toprule
\multirow{3}{*}{Objective Set} & \multirow{3}{*}{Method} & \multicolumn{10}{c}{Representation \& Predictive Performance} & \multicolumn{1}{c}{Concept Comp.} \\
\cmidrule(lr){3-12} \cmidrule(lr){13-13}
& & \multicolumn{3}{c}{Spearman $\uparrow$} & \multicolumn{2}{c}{V-score $\uparrow$} & \multicolumn{3}{c}{MAE$^{\circ}$ $\downarrow$} & \multicolumn{2}{c}{Acc. $\uparrow$} & \multicolumn{1}{c}{Sim. $\uparrow$} \\
\cmidrule(lr){3-5}\cmidrule(lr){6-7}\cmidrule(lr){8-10}\cmidrule(lr){11-12}\cmidrule(lr){13-13}
& & az & el & rot & cat & id & az & el & rot & cat & id & $\oplus$ and $\ominus$\\
\midrule
\multirow{2}{*}{Objective Set 1}

& Default & 0.95 & 0.87 & 0.85 & 0.99 & 0.91 & 0.08 & 0.08 & 0.09 & 0.99 & 0.97 & 0.95 \\
& \textbf{Continual}  & 0.92 & 0.81 & 0.84 & 0.82   & 0.93   & 0.11 & 0.11 & 0.12 & 0.94 & 0.87 & 0.72   \\
\bottomrule
\end{tabular}
\end{table}

\subsection{Analysis of Eigenvector-to-Concept Assignment}
\label{sec:appendix_assignment}

A key question regarding our framework is how each eigenvector $v_m$ is mapped to a specific task concept $C_m$. Our assignment is based on variance ranking (i.e., the top-$M$ eigenvectors are assigned to the $M$ objectives). However, the specific mapping (e.g., whether $v_0$ is assigned to \textit{azimuth} or \textit{category}) is arbitrary.

We hypothesize that this mapping is arbitrary because our training dynamic forces the alignment. The loss (as shown in Eq. 8 in the main paper) is computed after the projection. The gradient for a specific concept, $\mathcal{L}_m$, only backpropagates through its corresponding projected features, $f^{\text{proj}, m}$, which lie on the axis $v_m$. This training dynamic forces the encoder to learn to map all variance related to concept $C_m$ onto whichever axis $v_m$ it has been assigned.

To prove this, we ran a new experiment where we \textbf{randomly} shuffled the eigenvector-to-concept assignments and retrained the model from scratch. We tested the default (variance-ranked) assignment and two random shuffles (Default=$\{az,el,rot,cat,id\}$, Shuffle 1 = $\{cat,id,az,el,rot\}$, Shuffle 2 = $\{cat,id,az,el,rot\}$, assigned in order). The results are presented in Table~\ref{tab:shuffled_assignment}.

\begin{table}[h]
\centering
\small
\caption{Comparison of model performance with the default variance-ranked assignment versus randomly shuffled assignments. All models are trained on Objective Set 1. Arrows indicate whether higher ($\uparrow$) or lower ($\downarrow$) values are better.}
\label{tab:shuffled_assignment}
\footnotesize 
\setlength{\tabcolsep}{3.2pt}
\begin{tabular}{ll ccc cc ccc cc c}
\toprule
\multirow{3}{*}{Objective Set} & \multirow{3}{*}{Assignment} & \multicolumn{10}{c}{Representation \& Predictive Performance} & \multicolumn{1}{c}{Concept Comp.} \\
\cmidrule(lr){3-12} \cmidrule(lr){13-13}
& & \multicolumn{3}{c}{Spearman $\uparrow$} & \multicolumn{2}{c}{V-score $\uparrow$} & \multicolumn{3}{c}{MAE$^{\circ}$ $\downarrow$} & \multicolumn{2}{c}{Acc. $\uparrow$} & \multicolumn{1}{c}{Sim. $\uparrow$} \\
\cmidrule(lr){3-5}\cmidrule(lr){6-7}\cmidrule(lr){8-10}\cmidrule(lr){11-12}\cmidrule(lr){13-13}
& & az & el & rot & cat & id & az & el & rot & cat & id & $\oplus$ and $\ominus$\\
\midrule
\multirow{3}{*}{Objective Set 1}
& Default & 0.95 & 0.87 & 0.85 & 0.99 & 0.91 & 0.08 & 0.08 & 0.09 & 0.99 & 0.97 & 0.95 \\
& \textbf{Shuffle 1} & 0.95 & 0.87 & 0.85 & 0.98 & 0.95 & 0.08 & 0.08 & 0.09 & 0.99 & 0.91 & 0.87 \\
& \textbf{Shuffle 2} & 0.95 & 0.87 & 0.85 & 0.98 & 0.94 & 0.08 & 0.08 & 0.09 & 0.99 & 0.91 & 0.88 \\
\bottomrule
\end{tabular}
\end{table}

As shown in the table, the final performance is not negatively impacted by the random shuffling. This confirms that the model is not relying on a pre-existing alignment of variance; rather, our method \textit{creates} this alignment during training.


\subsection{Numerical Stability of Covariance Estimation}
\label{sec:appendix_stability}

A critical component of our framework is the eigendecomposition of the latent feature covariance matrix (as described in the main paper, Equation 3). The stability of this estimation is crucial for a stable orthogonal basis.

A naive approach would be to estimate the covariance matrix on each training mini-batch (e.g., $B_{\text{train}}=256$). However, this introduces significant instability. Our latent dimension is $D=2048$, and a typical training batch is much smaller than the dimensionality ($B_{\text{train}} \ll D$). This is a classic ``small $n$, large $p$'' problem, where estimating a $D \times D$ covariance matrix from $n$ samples is statistically ill-posed.  The resulting matrix is noisy and singular, and its eigenvectors are unstable, preventing the model from converging.

To solve this, we decouple the training batch size ($B_{\text{train}}$) from the covariance estimation set size ($B_{\text{cov}}$).

Our final methodology is as follows:
\begin{enumerate}
    \item At the beginning of each training epoch, we first compute latent features $\mathcal{F}$ for a large, stable subset of the training data, defined by $B_{\text{cov}}$.
    \item We compute the covariance matrix and its eigenvectors $V_M$ on this large, stable batch $\mathcal{F}$.
    \item For all training mini-batches (of size $B_{\text{train}}$) within that epoch, we \textbf{freeze} this orthogonal basis $V_M$.
\end{enumerate}

To determine how large $B_{\text{cov}}$ must be, we ran an ablation study varying its size as a percentage of the total training set. The results are shown in Table~\ref{tab:batch_size_ablation}. The performance is remarkably stable. Even when using only 10\% of the training data for the covariance estimation, the performance remains high. This confirms that our method is not only numerically stable, as a stable basis can be estimated from a small subset of the data.

\begin{table}[htbp] 
\centering
\small
\caption{Ablation study on the effect of the covariance estimation set size ($B_{\text{cov}}$), measured as a percentage of the total training set. All models are trained on Objective Set 1. Arrows indicate whether higher ($\uparrow$) or lower ($\downarrow$) values are better.}
\label{tab:batch_size_ablation}
\footnotesize 
\setlength{\tabcolsep}{3.2pt}
\begin{tabular}{ll ccc cc ccc cc c}
\toprule
\multirow{3}{*}{Objective Set} & \multirow{3}{*}{$B_{\text{cov}}$ (\% of Train)} & \multicolumn{10}{c}{Representation \& Predictive Performance} & \multicolumn{1}{c}{Concept Comp.} \\
\cmidrule(lr){3-12} \cmidrule(lr){13-13}
& & \multicolumn{3}{c}{Spearman $\uparrow$} & \multicolumn{2}{c}{V-score $\uparrow$} & \multicolumn{3}{c}{MAE$^{\circ}$ $\downarrow$} & \multicolumn{2}{c}{Acc. $\uparrow$} & \multicolumn{1}{c}{Sim. $\uparrow$} \\
\cmidrule(lr){3-5}\cmidrule(lr){6-7}\cmidrule(lr){8-10}\cmidrule(lr){11-12}\cmidrule(lr){13-13}
& & az & el & rot & cat & id & az & el & rot & cat & id & $\oplus$ and $\ominus$\\
\midrule
\multirow{4}{*}{Objective Set 1}
& 10\% & 0.96 & 0.87 & 0.85 & 0.99 & 0.94 & 0.08 & 0.08 & 0.09 & 0.99 & 0.97 & 0.94 \\
& 20\% & 0.96 & 0.87 & 0.87 & 0.98 & 0.95 & 0.08 & 0.08 & 0.09 & 0.99 & 0.97 & 0.95 \\
& 50\%  & 0.95 & 0.87 & 0.85 & 0.98 & 0.93 & 0.08 & 0.08 & 0.09 & 0.98 & 0.95 & 0.97 \\
& 100\%  (Default) & 0.95 & 0.87 & 0.85 & 0.99 & 0.91 & 0.08 & 0.08 & 0.09 & 0.99 & 0.97 & 0.95 \\
\bottomrule
\end{tabular}
\end{table}

\subsection{Robustness to Correlated or Redundant Tasks}
\label{sec:appendix_redundant}

A crucial question is whether our framework's strict orthogonality assumption is suboptimal or even detrimental when tasks are highly correlated or redundant. The concern is that enforcing orthogonality may "cost" the model, hurting performance on related tasks that could otherwise benefit from a shared representation.

To test this, we designed a new experiment where we explicitly introduced task redundancy. We took all 5 objectives from Objective Set 1 (az, el, rot, cat, id) and duplicated them, creating a new set of 5 "copy" tasks.

We then trained a new model on all 10 tasks simultaneously, assigning each of the 10 tasks to its own unique, orthogonal axis.

The results are presented in Table~\ref{tab:redundant_tasks}. We compare the performance of the default model (trained on 5 tasks) against the new model (trained on 10 tasks). The key finding is that the performance on the \textbf{original 5 tasks} remains stable, with no negative impact from the addition of the 5 redundant tasks. This suggests that our framework is robust to task redundancy and that the "cost" of enforcing orthogonality, even for highly correlated tasks, is not detrimental.

\begin{table}[htbp]
\centering
\small
\caption{Robustness to redundant tasks. We compare the default 5-task model against a model trained on 10 tasks (the 5 original + 5 duplicates). Performance on the original 5 tasks is not impacted, and the model successfully learns the new redundant tasks. Arrows indicate whether higher ($\uparrow$) or lower ($\downarrow$) values are better. (Note: The composition of the redundant task is reconstructed by 10 eigenvectors.)}
\label{tab:redundant_tasks}
\footnotesize 
\setlength{\tabcolsep}{3.2pt}
\begin{tabular}{ll ccc cc ccc cc c}
\toprule
\multirow{3}{*}{Objective Set} & \multirow{3}{*}{Method} & \multicolumn{10}{c}{Representation \& Predictive Performance} & \multicolumn{1}{c}{Concept Comp.} \\
\cmidrule(lr){3-12} \cmidrule(lr){13-13}
& & \multicolumn{3}{c}{Spearman $\uparrow$} & \multicolumn{2}{c}{V-score $\uparrow$} & \multicolumn{3}{c}{MAE$^{\circ}$ $\downarrow$} & \multicolumn{2}{c}{Acc. $\uparrow$} & \multicolumn{1}{c}{Sim. $\uparrow$} \\
\cmidrule(lr){3-5}\cmidrule(lr){6-7}\cmidrule(lr){8-10}\cmidrule(lr){11-12}\cmidrule(lr){13-13}
& & az & el & rot & cat & id & az & el & rot & cat & id & $\oplus$ and $\ominus$\\
\midrule
\multirow{3}{*}{Objective Set 1}
& Default & 0.95 & 0.87 & 0.85 & 0.99 & 0.91 & 0.08 & 0.08 & 0.09 & 0.99 & 0.97 & 0.95 \\
& \textbf{Original 5} & 0.94 & 0.85 & 0.84 & 0.98 & 0.83 & 0.09 & 0.09 & 0.10 & 0.99 & 0.99 & 0.91 \\
& \textbf{Redundant 5} & 0.85 & 0.84 & 0.84 & 0.98 & 0.83 & 0.09 & 0.09 & 0.10 & 0.97 & 0.94 & 0.91 \\
\bottomrule
\end{tabular}
\end{table}

\end{document}